\documentclass[a4paper,11pt,twocolumn,twoside]{article}
\usepackage[dvips]{graphicx}
\usepackage{sepln_en}
\usepackage{url}
\usepackage[utf8]{inputenc}
\usepackage{booktabs}
\usepackage{fullname}
\usepackage{mdframed}
\usepackage{float}
\usepackage{tikz}
\usepackage{pgfplots}
\pgfplotsset{compat=1.18}
\usepackage{ltablex} 
\usepackage{multirow} 
\usepackage{pifont}
\usepackage[gen]{eurosym}
\usepackage{placeins}
\usepackage{covington} 

\usepackage[dvipsnames]{xcolor}
\usetikzlibrary{patterns, backgrounds}

\input epsf

\title{BOE-XSUM: Extreme Summarization in Clear Language of Spanish Legal Decrees and Notifications}

\author {\textbf{Andrés Fernández García$^{1,*}$}, \textbf{Javier de la Rosa$^{2,*}$},
\textbf{Julio Gonzalo$^1$}, \textbf{Roser Morante$^1$},\\ \textbf{Enrique Amigó$^1$}, \textbf{Alejandro Benito-Santos$^1$}, \textbf{Jorge Carrillo-de-Albornoz$^1$}\\ ,\textbf{Víctor Fresno$^1$}, \textbf{Adrian Ghajari$^1$}, \textbf{Guillermo Marco$^1$}, \textbf{Laura Plaza$^1$}, \textbf{Eva Sánchez Salido$^1$}\\.
\\
$^1$Universidad Nacional de Educación a Distancia, Spain\\
$^2$The National Library of Norway, Norway\\
\small
    \textbf{Correspondence:} {\texttt{nandezgarcia@gmail.com}}
\\
}

\seplntranstitle{BOE-XSUM: Resúmenes Extremos en Lenguaje Claro de Decretos y Notificaciones Legales en Español}

\seplnclave{Resumen extremo, texto legal, modelos generativos, recursos de evaluación.}

\seplnresumen{
La capacidad de resumir documentos largos de forma concisa es cada vez más importante en la vida cotidiana debido a la sobrecarga de información, pero existe una notable escasez de este tipo de resúmenes para documentos en español en general, y en el ámbito jurídico en particular. En este trabajo, presentamos BOE-XSUM, un conjunto de datos de 3648 resúmenes extremadamente breves en lenguaje claro creados a partir de las entradas del Boletín Oficial del Estado (BOE). El conjunto de datos contiene tanto los resúmenes como los textos originales etiquetados con el tipo de documento. Además, presentamos los resultados de experimentar en modo de fine-tuning y de zero-shot con modelos generativos. Nuestros resultados indican que los modelos generativos supervisados mediante fine tuning funcionan significativamente mejor que los modelos generativos en modo no supervisado, incluso siendo modelos más pequeños. El mejor modelo con finetuning de nuestra experimentación, BERTIN GPT-J 6B (precisión de 32 bits), obtiene resultados un 24\% mejores que el mejor modelo no supervisado, DeepSeek-R1 (41,6\% vs 33,5\%). 
}

\seplnkey{Extreme-summarization, legal texts, generative models, evaluation resources.}

\seplnabstract{The ability to summarize long documents succinctly is increasingly important in daily life due to information overload, yet there is a notable lack of such summaries for Spanish documents in general, and in the legal domain in particular. In this work, we present BOE-XSUM, a curated dataset comprising 3,648 concise, plain-language summaries of documents sourced from Spain’s ``Boletín Oficial del Estado" (BOE), the State Official Gazette. Each entry in the dataset includes a short summary, the original text, and its document type label. We evaluate the performance of medium-sized large language models (LLMs) fine-tuned on BOE-XSUM, comparing them to general-purpose generative models in a zero-shot setting. Results show that fine-tuned models significantly outperform their non-specialized counterparts. Notably, the best-performing model—BERTIN GPT-J 6B (32-bit precision)—achieves a 24\% performance gain over the top zero-shot model, DeepSeek-R1 (accuracies of 41.6\% vs. 33.5\%).}

\firstpageno{1}

\begin{document}


\setlength\titlebox{22cm} 

\label{firstpage} \maketitle

\renewcommand{\thefootnote}{\fnsymbol{footnote}}
\footnotetext[1]{These authors contributed equally to this work.}
\renewcommand{\thefootnote}{\arabic{footnote}}  

%

\section{Introduction}
Among the many capabilities of large language models (LLMs), summarization stands out as one of the most widely used and valued by end users \cite{Budhiraja2024,brachman2025current}. Recent advances in LLMs have made it possible to summarize extremely long documents with remarkable accuracy and coherence. These breakthroughs have been driven in large part by the availability of large-scale, high-quality summarization datasets—particularly in English. However, this progress has not been evenly distributed across languages. Spanish, despite being the second most widely spoken native language in the world with over 485 million speakers \cite{ethnologue2023mostspoken}, remains significantly under-resourced in Natural Language Processing (NLP) \cite{Conde:2024}. This scarcity is especially pronounced in summarization tasks: a search on Hugging Face reveals 963 summarization datasets in English,\footnote{\url{https://huggingface.co/datasets?task_categories=task_categories:summarization&language=language:en&sort=trending}, accessed on April 1st, 2025.} compared to only 75 in Spanish.\footnote{\url{https://huggingface.co/datasets?task_categories=task_categories:summarization&language=language:es&sort=trending}, accessed on April 1st, 2025.} Moreover, domain-specific resources—such as those tailored for administrative or legal document summarization—are virtually non-existent in Spanish, further limiting the development and applicability of LLMs in these critical areas.



In this work, we introduce BOE-XSUM, a curated dataset comprising 3,648 clear and extremely concise summaries of entries from Spain’s ``Boletín Oficial del Estado" (BOE)\footnote{\url{https://www.boe.es/}}, the country’s State Official Gazette, along with their original texts. This gazette serves as the primary platform for disseminating legislative decrees, legal notifications, and various official documents, predominantly originating from the national government, but including contributions from regional and local authorities. This is the first dataset of its kind in Spanish and is characterized by two key features: first, it addresses the important challenge of adapting complex domain-specific language, such as legal or administrative texts, into clear, everyday language; and second, it provides extremely concise summaries that are manually curated and verified. 

The extreme summaries are built upon social media posts by a Spanish journalist specialized in the analysis and treatment of public information.\footnote{Eva Belmonte is a journalist and co-director of Civio Foundation (see \url{https://civio.es/}) an independent, non-profit organization that monitors public authorities through data journalism, and works on three lines of action: journalism, public advocacy and transparency services for public administrations. Belmonte is part of a multidisciplinary group of people who work to improve the democratic quality in Spain. She publishes concise daily summaries of the BOE in X (see \url{https://x.com/evabelmonte}).} This is a socially relevant task for the control of governments, whether they are central, autonomous, or municipalities, and other public agencies such as the Constitutional and the Supreme Court. For example, in the context of the floods that devastated several regions in Spain in late 2024,\footnote{\url{https://en.wikipedia.org/wiki/2024_Spanish_floods}.} these posts contributed to inform the population about how central and regional governments were managing public resources to alleviate the effects of the disaster.  

By compiling and making available the BOE-XSUM dataset, we aim to fill the gap in NLP resources for Spanish, offering a valuable resource for both academic research and practical applications. In addition, we present experiments with generative language models. This allows us to answer the following research question: to what extent can generative language models produce extreme summaries of legal texts, not only capturing their complex meaning but also adapting them into clear language in a manner comparable to that of a human expert?

Our contribution is twofold. First, we release a new publicly available dataset of extreme summaries in Spanish within the legal and administrative domain.\footnote{\url{https://huggingface.co/datasets/bertin-project/BOE-XSUM}} Second, we conduct experiments with generative text models, revealing that current systems still struggle to produce high-quality extreme summaries. However, a qualitative analysis shows that in some cases, the generated summaries resemble those written by humans. 


This paper is organized as follows. In Section~\ref{s:relatedwork}  we present the related work. In Section~\ref{s:dataset} we describe the BOE-XSUM dataset and provide a qualitative analysis of its extreme summaries. Section~\ref{s:experiments} focuses on the experiments, the results of which are discussed in Section~\ref{s:discussion}. Finally, the conclusions are presented in Section~\ref{s:conclusions}.

\section{Related work}
\label{s:relatedwork}

In general, there are two main approaches to summarization: extractive and abstractive summarization \cite{cajueiro2023comprehensivereviewautomatictext}. Within the extractive works, we can find those that compute frequencies, such as those based on spatial vectors \cite{ModernInformationRetrieval,BELWAL2021102536}, matrix factorization \cite{10.1145/383952.383955}, graphs \cite{mihalcea-tarau-2004-textrank}, topics \cite{Haghighi2009ExploringCM} and neural word embeddings \cite{inproceedingsvectors}. In addition to frequency-based methods, there are also approaches based on heuristics \cite{Edmundson1969NewMI}, \cite{10.1145/1980022.1980170}, linguistics \cite{10.1145/366532.366545}, \cite{Mohamed2016AutomaticTS}, supervised machine learning \cite{MAO2019173}, and reinforcement learning \cite{ryang-abekawa-2012-framework,hyun2022generatingmultiplelengthsummariesreinforcement}. Extreme summarization is a form of single-document summarization that aims to generate highly concise summaries—often a single sentence—that capture the core meaning of the source text \cite{narayan2018don,cachola2020tldr}. Unlike extractive methods, it requires abstractive generation to synthesize information from the input, making it particularly useful in domains like scientific literature and news media \cite{mao2022citesum}.

There exist datasets of summaries in many languages such as English \cite{grusky-etal-2018-newsroom}, Spanish and Catalan \cite{segarra-soriano-etal-2022-dacsa}, Indonesian \cite{koto-etal-2020-liputan6}, and Bengali \cite{khan-etal-2023-banglachq} among others. And while there are multilingual datasets available, such as MLSUM \cite{scialom2020mlsum}, WikiLingua \cite{ladhak2020wikilingua}, EUR-Lex-Sum \cite{aumiller2022eurlexsum}, XL-SUM \cite{hasan2021xlsum} and HumSet \cite{fekih2022humset}, they usually share certain characteristics that are not optimal for their use in training generative models. For instance, summaries may be provided as titles preceding the text, or they might conclude with a link to the full text, making it very easy for language models to match the original text and its summary. Such features, particularly given the public nature of these texts, enhance the effectiveness of generative models, which, in the end, perform a memorization task, instead of a meaning abstraction task.  

As for extreme summarization datasets in Spanish, the availability of resources is severely limited, with only one available dataset, {NoticIA} \cite{garcíaferrero2024noticiaclickbaitarticlesummarization}, which consists of 850 Spanish news articles with clickbait headlines, each paired with a human-written, single-sentence generative summary. This dataset assesses the ability of models to understand and summarize texts, addressing the challenge of interpreting complex information generated by clickbait headlines. Finding datasets of extreme summaries in languages other than English or Chinese is really difficult \cite{Narayan2018DontGM,sotudeh2021tldr9largescaleresource,cachola2020tldrextremesummarizationscientific}.

The challenge of summarizing complex content is particularly evident when translating from highly specialized technical language—such as legal or scientific discourse—into more accessible forms. This issue has been widely explored in the context of science communication and plain language movements. Studies in popular science writing have shown how lexical and structural simplification strategies can effectively bring technical content closer to general audiences, while still preserving meaning and nuance \cite{montalt_gonzalezdavies}. Similarly, the field of legal communication has highlighted the importance of translating dense, technical legal language into clear and formal language for broader accessibility \cite{tiersma_plainlanguage}. These linguistic transformations not only benefit public understanding but are also central to tasks such as summarization, where preserving intent while changing register is essential. In this sense, research on genre translation—from specialized to general discourse—offers valuable frameworks for evaluating the capacity of language models to handle extreme summarization tasks in real-world scenarios \cite{baramtsabari_lewenstein}.

\section{Dataset}
\label{s:dataset}

In this section we provide details about the dataset creation: data sources, data extraction, data cleaning, and annotations. 

\subsection{Sources}

The BOE-XSUM dataset originates from the social media posts made by a Spanish journalist who performs daily reviews of the BOE. She selects the articles that considers of greatest social interest and summarizes them in posts that also include a the link to the PDF version of the specific BOE article being summarized. Depending on the relevance or complexity of the content, the journalist writes one or several short posts summarizing the essence of the resolutions and orders in the article of the BOE. In the case of crucial matters, she elaborates a detailed long post on an external website, which is subsequently linked to the initial post. Each summary in the dataset has two versions: the original social media post written by the journalist, and an edited version crafted for the extreme summarization task. These edited versions have been refined to ensure accurate representation of the BOE content and to meet the standards of clear language summarization. All experiments presented in this work have been conducted  using the edited summaries.

We initially collected over 4,500 social media posts. After manual review, we discarded those that lacked a direct link to a BOE article, were not clearly related to the BOE content, or were too subjective. Detailed annotation guidelines are available in Appendix \ref{app:guidelines}. The final dataset consists of 3,648 BOE entries, each paired with an original post and its corresponding edited summary. The BOE articles were also annotated by category to enhance their usability for both summarization and classification tasks.


\subsection{Data extraction}


The next step was to classify the links included in the social media posts into two categories. Given the occasional use of external links, our analysis focused on distinguishing and separating between two main categories of links: those linking to the journalist external website, which contains articles related to the BOE, and those linking directly to BOE documents. This differentiation allowed us to capture both the concise summaries inherent in her posts and, where appropriate, extended summaries along with direct links to BOE documents. In the case of links to BOE documents, mostly in PDF format, a conversion process was employed to facilitate their download in plaint text format, preserving the identifier used originally in the BOE article.\footnote{For example, from PDF link from the BOE as \url{https://boe.es/boe/dias/2024/03/28/pdfs/BOE-A-2024-6273.pdf}, we  first take its identifier "BOE-A-2024-6273", \url{https://boe.es/diario_boe/txt.php?id=BOE-A-2024-6273}, and then use it to generate the official URL that exposes the content in text format, which we subsequently use to download the content of the articles in plain text for our dataset.}

\subsection{Manual data review}\label{data_review}

After a preliminary analysis of the dataset, a decision was made to develop a tool for the visualization, tagging, and editing of the dataset, as depicted in Figure \ref{fig:server_editor} included in Appendix~\ref{ap:editor}. The main purpose of the tool was to verify the integrity of the data visualization and being able to modify the data if there were any errors. The review of the data was carried out with this tool. We proceeded to verify, one by one, the correspondence between the BOE text and the tweet containing the associated summary. When we identified any inconsistencies, such as a summary that did not match the content of the BOE or that was poorly contextualized, we made the necessary corrections to ensure that the summary accurately reflected the information contained in the official text. The complete annotation guidelines are described in Appendix \ref{app:guidelines}. 

We also edited a number of summaries to ensure that they better reflected the content of the original BOE document and agreed to our guidelines, improving clarity and accuracy. Both the original and edited versions are part of the dataset. Importantly, all training and evaluation experiments described in this work have been carried out using the edited versions only. This ensures consistency, eliminates ambiguity, and aligns with the goal of producing high-quality summaries in clear language. In Appendix \ref{ap:edited} we show two examples of how the summaries were edited.


To estimate the extent of the editing process, we conducted a similarity analysis based on cosine distance between original and edited versions.\footnote{We extracted embedding vectors using the original multilingual BERT.} Table~\ref{t:similarity-distribution} in Appendix~\ref{ap:additional-info-dataset} shows that a third of the dataset (1,154 posts) had a similarity above 90\%, while approximately 160 posts had a similarity below 10\%. This variation reflects the range of interventions applied--from minor adjustments to full rewrites--to ensure that the summaries do not only provide a consistent view, but also faithfully represent the original BOE content.


Moreover, in a test with two male participants aged between 30 and 45 years, both holding university degrees, summaries edited according to the guidelines were chosen 189 out of 200 times (94.5\%), 95\% CI [90.42\%, 96.9\%], significantly above chance (two-tailed binomial test, $p < 0.001$). 




\subsection{Categorization}

Each entry in the dataset includes both the original and edited summaries, along with a label indicating the type of BOE article. All categories are disjoint and the labeling of articles follows two main criteria:

{\small

\begin{enumerate}
    \item \textbf{Explicit Mention in the BOE article:} If the name of a category is explicitly stated in the BOE article, that label is assigned directly. For example:
    \begin{itemize}
        \item Articles from \texttt{TRIBUNAL\_CONSTITUCIONAL} are labeled as \emph{Constitutional Court}.
        \item Articles from \texttt{CONVENIOS} are labeled as \emph{Agreement}.
    \end{itemize}
    Some categories are formed by combining multiple clearly defined types. For instance, articles related to awards and medals are grouped under the \texttt{PREMIOS\_Y\_MEDALLAS} category.
    
    \item \textbf{Frequency-Based Filtering:} Categories with a very low number of articles are grouped into a general category named \texttt{OTROS\_ANUNCIOS} (\textit{Other announcements}). One exception is the \texttt{BANCO\_DE\_ESPAÑA} (\textit{Bank of Spain}) category, which is retained in the dataset due to its relevance, even though it contains relatively few articles.
\end{enumerate}
}

A list of all categories with illustrative examples is provided in Table~\ref{t:categories-examples} in Appendix~\ref{ap:catgories}. The distribution of categories across the dataset partitions is shown in Table~\ref{t:distribution-categories}.

As presented in Table~\ref{tab:dataset_columns}, the dataset contains several columns. Among the most important are: BOE article texts (text), Original posts (summaries), Edited versions of the posts (edited summaries).

\subsection{Content Analysis}
\label{s:analysis-data}

Among the BOE documents, there exists a variety of contents, including straightforward articles such as the appointments of ambassadors or designation of official positions. These particular entries, by virtue of their inherently concise nature, offer a less complex summarization task. 

Conversely, the dataset also encompasses BOE articles of a more intricate and voluminous nature, including but not limited to decisions from the Constitutional Court, Supreme Court rulings, and various agreements. Such documents are characterized by their extensive length, often spanning thousands of words. To illustrate the complexity and scale of these articles, we show a simplified  example, with only the  introductory and concluding segments of this representative article:\footnote{See English translations in Appendix~\ref{app:english_example}.}


\begin{mdframed}
\footnotesize
El Real Decreto-ley 6/2012, de 9 de marzo, de medidas urgentes de protección de deudores hipotecarios sin recursos, establece una serie de mecanismos conducentes a permitir la reestructuración de la deuda hipotecaria de quienes padecen extraordinarias dificultades para atender su pago.
A tal fin, al citado Real Decreto-ley se incorporó un código de buenas prácticas al que podrán adherirse las entidades y cuyo seguimiento será supervisado por una comisión de control, cuya composición ha sido modificada por el artículo 6 de la Ley 1/2013, de 14 de mayo, de medidas para ...
... – Caja Rural San Jaime de Alquerías Niño Perdido, S. Coop. de Crédito V.
– Caja Rural San José de Almassora, S. Coop. de Crédito. V.
– Caja Rural San José de Burriana, S. Coop. de Crédito V.
– Caja Rural San José de Nules, S. Coop. de Crédito V.
– Caja Rural San Roque de Almenara, S. Coop. de Crédito V.
– Colonya-Caixa D’estalvis de Pollença.
– Liberbank, S. A.
– Publicredit, S. L.
– UNOE Bank, S. A.
\end{mdframed}



The original summary for this BOE article as written by the journalist is as follows:

\begin{mdframed}
\footnotesize
    Adhesiones a 2 códigos de buenas prácticas en hipotecas  Has ahora, poco efectivos. Porque legislar no, ¿verdad? \#BOE
\end{mdframed}

The resulting edited summary:

\begin{mdframed}
\footnotesize
    Lista de adhesiones de bancos a los  códigos de buenas prácticas para reforzar la protección a los deudores hipotecarios, reestructuración de deuda y alquiler social
\end{mdframed}

\begin{figure}[H]
    \small
    \centering
    \begin{tikzpicture}
\begin{axis}[
    width=7.5cm,
    height=4cm,
    xlabel={Words per document},
    ylabel={BOE documents},
    xmin=0, xmax=14.5,
    ymin=0, ymax=42,
    xtick={0, 3.5, 7, 10.5, 14},
    xticklabels={25K, 50K, 100K, 150K, 200K},
    xticklabel style={font=\footnotesize},
    yticklabel style={font=\footnotesize},
    grid=none,
    axis lines=left,
    bar width=8000,
    area style,
]

\addplot+[
    ybar interval,
    fill=blue!70,
    draw=blue!70,
    mark=none
] coordinates {
    (0, 39)
    (1, 13)
    (2, 12)
    (3, 9)
    (4, 9)
    (5, 5)
    (6, 1)
    (7, 1)
    (8, 4)
    (9, 0)
    (10, 3)
    (11, 0)
    (12, 0)
    (13, 1)
    (14, 1)
};

\end{axis}
\end{tikzpicture}
    \caption{Histogram showing the distribution of BOE documents by word count for BOE documents with at least 25,000 words. \label{fig:histogram_less_1000}}
    
\end{figure}
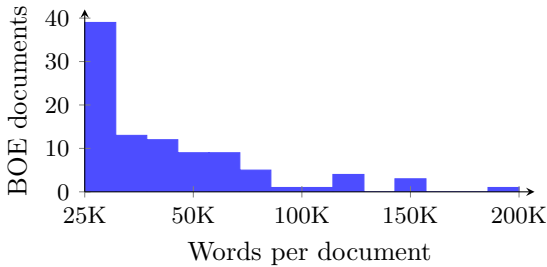

As Table~\ref{fig:histogram_less_1000} shows, the dataset contains a total of 3,648 BOE texts, with a total 13,304,989 space-delimited words. The average number of words per document is 3,396, while the summaries average a total of 17 words, which gives us a 0.005\% compression rate from the original text to the summary. More than 64\% of BOE documents have less than 1,000 words and only 2.65\% have more than 25,000 words. A histogram for BOE documents with at least 25,000 words can be seen in Figure \ref{fig:histogram_less_1000}.\footnote{This is due to the inclusion of complete external texts in the BOE document itself, such as Royal Decrees and Laws. See also Table \ref{t:similarity-distribution} in Appendix \ref{ap:additional-info-dataset} for
 the percentages of texts by ranges of increasing size.}

The dataset was divided into three splits: train, development, and test. Initially, we adhered to an 80/10/10 distribution, but adjustments were made to ensure that the annotated categories were well-balanced across the splits. The distribution of categories per split is provided in Table \ref{t:distribution-categories} of Appendix~\ref{ap:additional-info-dataset}.
\section{Experiments}
\label{s:experiments}

We address the task of extreme summarization using BOE-XSUM, which requires deep textual comprehension to produce highly condensed summaries of no more than 280 characters. This length constraint was chosen to parallel the character limit typical of the social media platform on which the original summaries were written. The degree of compression varies significantly across documents; some entries in the BOE corpus are relatively brief, while others are substantially longer, as detailed in the analysis section. We evaluate language models under two settings: fine-tuning and zero-shot prompting.

Although the dataset also includes the original posts for reference, all experiments are conducted using the edited summaries produced through the manual curation process described in Section \ref{data_review}. This ensures a consistent and contextually accurate benchmark for the extreme summarization task.

\subsection{Evaluation Metrics}

The evaluation of LLMs performed using common metrics in the literature: BLEU \cite{papineni-etal-2002-bleu}, ROUGE \cite{lin-2004-rouge}, METEOR \cite{10.5555/1626355.1626389},  and BERTScore \cite{zhang2020bertscoreevaluatingtextgeneration}. BLEU focuses on n-gram overlap with reference texts and includes a brevity penalty to discourage overly short outputs. METEOR emphasizes unigram precision and recall with added flexibility through synonym and stem matching, and includes penalties for fragmented matches. ROUGE measures quality using n-gram overlap and the longest common subsequence, with variants tailored to different summarization aspects. BERTScore leverages contextual embeddings from models like BERT to assess semantic similarity via cosine similarity, enabling a more nuanced comparison beyond surface-level word matching.

\begin{table*}[htb]
\centering
\small
\footnotesize
\begin{tabular}{lcccccc}
\textbf{Model} & \textbf{Precision} & \textbf{Layers} & \textbf{BLEU} & \textbf{METEOR} & \textbf{ROUGE} & \textbf{BERTScore} \\
\midrule
\multirow{6}{*}{BOLETIN} & 4-bit & 13 & 0.050 & 0.258 & 0.281 & 0.262 \\
 & 8-bit & 13 & 0.065 & 0.292 & 0.306 & 0.289 \\
 & 16-bit & 13 & 0.066 & 0.299 & 0.308 & 0.294 \\
 & 4-bit & 27 & 0.054 & 0.264 & 0.287 & 0.269 \\
 & 8-bit & 27 & 0.062 & 0.292 & 0.305 & 0.286 \\
 & 16-bit & 27 & 0.074 & 0.306 & 0.317 & 0.298 \\
 & 32-bit & All & \textbf{0.094} & \textbf{0.333} & \textbf{0.367} & \textbf{0.397} \\
\midrule
\multirow{6}{*}{BERTIN GPT-J 6B} & 4-bit & 13 & 0.062 & 0.274 & 0.289 & 0.264 \\
 & 8-bit & 13 & 0.057 & 0.295 & 0.304 & 0.287 \\
 & 16-bit & 13 & 0.061 & 0.306 & 0.314 & 0.299 \\
 & 4-bit & 27 & 0.068 & 0.278 & 0.293 & 0.267 \\
 & 8-bit & 27 & 0.056 & 0.295 & 0.305 & 0.291 \\
 & 16-bit & 27 & 0.057 & 0.305 & 0.313 &  0.300 \\
 & 32-bit & All & {\bf 0.109} & {\bf 0.365} & {\bf 0.393} &	{\bf 0.416} \\
\end{tabular}
\caption{Scores of the differet metrics for fine-tuned BOLETIN and BERTIN GPT-J 6B models evaluated on the test split. The models differ in precision (bit-depth) and the number of layers trained (13 or 27 for LoRA training, ``All" for full fine-tuning). The last row for each model corresponds to the result for a 32-bit model with all layers trained. Best scores in \textbf{bold}.}
\label{tab:result_finetuning}
\end{table*}

\subsection{Fine-tuning experiments}
We used BERTIN GPT-J 6B\footnote{\url{https://huggingface.co/bertin-project/bertin-gpt-j-6B}} \cite{BERTIN,bertin-gptj} as the base model. After downloading the entire BOE from 1988 to 2023, we continued pre-training this model for 3 epochs using the same parameters as the original configuration, resulting in a domain-adapted variant, BOLETIN.\footnote{\url{https://huggingface.co/bertin-project/BOLETIN}} With these two models in place, we conducted a grid search to train them under various configurations.

\begin{table}[h!]
\centering
\footnotesize
\begin{tabular}{ll}
\textbf{Parameter} & \textbf{Value} \\
\midrule
Batch size & 4 \\
Gradient accumulation steps & 4 \\
Warmup steps & 100 \\
Max steps & 600 \\
Learning rate & 2e-4 \\
\end{tabular}
\caption{Fine-tuning hyperparameters for all experiments.}
\vspace*{-4\baselineskip}
\label{tab:training-params}
\end{table}

To establish a performance baseline, we first fully fine-tuned both models using 32-bit precision. In parallel, we explored parameter-efficient fine-tuning using Low-Rank Adaptation (LoRA) \cite{hu2022lora}. LoRA introduces trainable low-rank matrices into key components of the model’s attention mechanism—specifically the layers responsible for projecting and integrating attention information—allowing the model to adapt to new tasks while updating only a small subset of parameters. This significantly reduces computational and memory requirements. Training was conducted for up to 600 steps using mixed-precision formats (4-, 8-, and 16-bit), as detailed in Table~\ref{tab:training-params}. Results in Table~\ref{tab:result_finetuning} illustrate the trade-offs between full and parameter-efficient fine-tuning across model variants.

Given the architectural constraints of these models, the input sequence is limited to a maximum of 2048 tokens. To accommodate summary generation within this limit, a portion of tokens must be reserved for both initiating the summary and for the model’s output. During training, we appended the marker ``\texttt{\#\#\# RESUMEN:}" (``\textit{\texttt{\#\#\# SUMMARY:}}") after each BOE article, followed by its corresponding summary. In cases where an article exceeded the token limit, we truncated the input as to accommodate for the task marker and allow sufficient space for model output during generation. For especially long documents, we ensured that the summary referred primarily to the initial portion of the article, minimizing the risk that truncation would remove critical information required for accurate summarization. However, these heuristics were not always sufficient as many BOE texts are long and complex. As a result, a non-negligible number of examples likely lacked critical information for the generation of accurate summaries. We did not systematically exclude or mark truncated samples, which means that some training instances may have introduced noise or incomplete context. This could partially explain the issues observed in some generated summaries, which were vague or abruptly cut. Future work should consider integrating models with larger context windows or using strategies such as sliding windows or hierarchical encoding to handle long legal texts more effectively.

Table \ref{tab:result_finetuning} provides the results of the fine-tuning experiments for both BERTIN GPT-J 6B and BOLETIN. Despite the high level of complexity of the task, most models achieve similar outcomes, with full-precision fine-tuning outperforming any LoRA configuration. We also noted that the number of trained layers is not predictive of performance (see Figure \ref{fig:comparative_boletin_bertin} in Appendix~\ref{ap:additional-results}).

\begin{table}[h!]
\centering
\footnotesize
\begin{tabular}{l|cccc}
 & \textbf{B} & \textbf{M} & \textbf{R} & \textbf{BS} \\
\hline
\textbf{BLEU} (B)      & 1.000  & 0.872  & 0.936  & 0.919 \\
\textbf{METEOR} (M)   & 0.872  & 1.000  & 0.967  & 0.934 \\
\textbf{ROUGE} (R)    & 0.936  & 0.967  & 1.000  & 0.991 \\
\textbf{BERTScore} (BS) & 0.919  & 0.934  & 0.991  & 1.000 \\
\end{tabular}
\caption{Pearson correlation matrix between automatic evaluation metrics.}
\label{fig:correlation}
\end{table}

Moreover, Table~\ref{fig:correlation} shows that all metrics are strongly correlated with each other, suggesting that they measure related aspects of the quality of the generated text. ROUGE and BERTScore show the highest correlation (0.991), indicating that they tend to vary together and reflect very similar characteristics of the content, such as coverage and semantics. BLEU has the lowest correlation with METEOR (0.872), but it is still high. This makes sense, as BLEU is more strict (based on exact n-gram matches), while METEOR is more flexible and semantically oriented. Overall, using these metrics together provides a coherent and consistent evaluation of model performance.



Interestingly, despite the additional pretraining of the BOLETÍN model on the full content of the BOE (1988–2023) to enhance its legal domain knowledge, this strategy did not lead to improved performance on the extreme summarization task in plain language. Our initial hypothesis was that continued pretraining would help the model better grasp the structure, terminology, and semantics of legal and administrative texts, thereby improving its ability to generate accurate summaries. While this approach may be beneficial for tasks that require formal or technical language, it appears to be less effective when the objective is to produce highly concise summaries in clear, accessible language. The specialized patterns reinforced during domain adaptation may have introduced a linguistic mismatch with the target style. This result underscores the importance of aligning not only the domain but also the linguistic register of the training data with the specific demands of the downstream task.
    
A deeper analysis revealed that the length constraint during training impacted negatively the generation of summaries. For example, we encountered incomplete summaries like: \textit{`Real Decreto que modifica el Real Decreto 369/1999, de 5 de marzo, sobre términos y...'}, instead of the expected summary: \textit{`Curas evangélicos que acrediten haber ejercido antes de 1999 cobrarán pensión de jubilación.'} Or cases like: \textit{`El CSD notifica a todos los interesados en el recurso del Real Madrid contra las modificaciones de los...'}, where the correct summary should have been: \textit{`El Real Madrid recurre las reformas de reglamento y estatutos sociales de la Liga de Fútbol.'}

\subsection{Prompting experiments}

For zero-shot experimentation, we selected a variety of open source and proprietary models of different sizes, some multilingual some tailored to Spanish content. Our goal was to analyze the performance of both large and small models.

In this setting, the importance of the prompt is crucial in determining the quality of response of a generative model. After some iterations and a limited number of manual trials, we settled on a prompt that showed promise of being precise and effective in the generation of summaries.

\begin{description}
    \item[Prompt:] \textit{Eres un experto generando resúmenes en lenguaje cotidiano a partir de documentos legales escritos en lenguaje formal. Quiero que me des un resumen en español de entre 15 y 22 palabras del siguiente texto. Recuerda, solo quiero que devuelvas el resumen, nada más. Devuelveme únicamente el resumen, solo  el resumen. A continuación te indico el texto que debes resumir: [\textsc{boe document}] }
\end{description}


\begin{table*}[htb]
\centering
\footnotesize
\begin{tabular}{lccccc}
\textbf{Model} & \textbf{BLEU} & \textbf{METEOR} & \textbf{ROUGE} & \textbf{BERTScore} & \textbf{Avg. Words} \\
\midrule
Gemma 2 27b & 0.041 & 0.221 & 0.245 & 0.266 & 12.02 \\
Gemma 2 9B & 0.042 & 0.211 & 0.238 & 0.255 & 14.93 \\
Gemma 2 2B & 0.041 & 0.194 & 0.222 & 0.248 & 16.53 \\
ChatGPT 4o & 0.042 & \underline{0.265} & \underline{0.314} & 0.327 & 16.65 \\
Llama 3 70b Instruct & \underline{0.044} & 0.227 & 0.269 & 0.285 & 18.75 \\
DeepSeek R1 & 0.041 & 0.251 & 0.307 & \underline{0.335} & 18.99 \\
Nemotron 70B & 0.035 & 0.183 & 0.249 & 0.205 & 19.93 \\
Llama 3.2 3B & 0.035 & 0.197 & 0.236 & 0.245 & 21.40 \\
Llama 3.2 1B  & 0.020 & 0.143 & 0.187 & 0.174 & 31.69 \\
Llama 3.1 70B & 0.029 & 0.229 & 0.238 & 0.245 & 31.69 \\
Llama 2 70B & 0.038 & 0.156 & 0.216 & 0.214 & 33.72 \\
Llama 3.1 8B  & 0.034 & 0.199 & 0.243 & 0.251 & 33.72 \\
Salamandra 7B Instruct & 0.013 & 0.113 & 0.171 & 0.141 & 66.24 \\
Solar 10.7B & 0.011 & 0.110 & 0.172 & 0.166 & 60.70 \\
Neural-Chat 7B & 0.011 & 0.125 & 0.193 & 0.187 & 57.77 \\
Mistral 7B & 0.015 & 0.107 & 0.176 & 0.158 & 73.65 \\
Starling 7B & 0.007 & 0.088 & 0.157 & 0.149 & 72.92 \\
Llava 7B & 0.007 & 0.080 & 0.144 & 0.126 & 88.48 \\
\midrule
BERTIN GPT-J 6B & {\bf 0.109} & {\bf 0.365} & {\bf 0.393} & {\bf 0.416} & 16.10 \\
\end{tabular}
\caption{Performance results of the generative models with prompts across different metrics. Added BERTIN GPT-J 6B for reference. Best scores in \textbf{bold}, second best \underline{underlined}.
\label{tab:resultados_modelos_prompt}}
\end{table*}

Table~\ref{tab:resultados_modelos_prompt} shows the results from the experiments using the edited summaries\footnote{See Table \ref{tab:baseline_sim_1} in Appendix \ref{ap:additional-results} for results obtained using the original unmodified original posts by the journalist, which are generally worse.}. On BOE-XSUM, DeepSeek R1 produced the best results with an BERTScore of 0.335, followed by the model ChatGPT 4o with a BERTScore of 0.327 and Llama 3 70b Instruct with a score of 0.285. Gemma 2 27B obtained a 0.266. It is remarkable that the model Gemma 2 9B produced better summaries than Llama 2 70B, given their difference in size.\footnote{See Table \ref{tab:example_1_gemma2vsLlama2} in Appendix \ref{ap:additional-results} for examples of Gemma 2 2B vs Llama 2 70B.} We also observed that Llama 2 70B was prone to generate much longer summaries, while ChatGPT 4o, Llama 3 70B Instruct, and Gemma 2 models produced shorter and more accurate summaries. The three Gemma 2 models produced the shortest summaries, but they got worse as the model got smaller. It is also striking that the average length of the summaries produced by the top-ranked models, ChatGPT 4o and DeepSeek R1, is close to that of those of the ground truth summaries (17 words), with an average of 16.68 and 18.99 words, respectively. Importantly, there seems to exist a strong negative correlation between the average number of words of generated summaries and their BERTScore values across models (Pearson's $r = -0.82$, $p < 0.001$), suggesting that as the length of a generated summary increases, its BERTScore against the ground truth tends to decrease significantly.

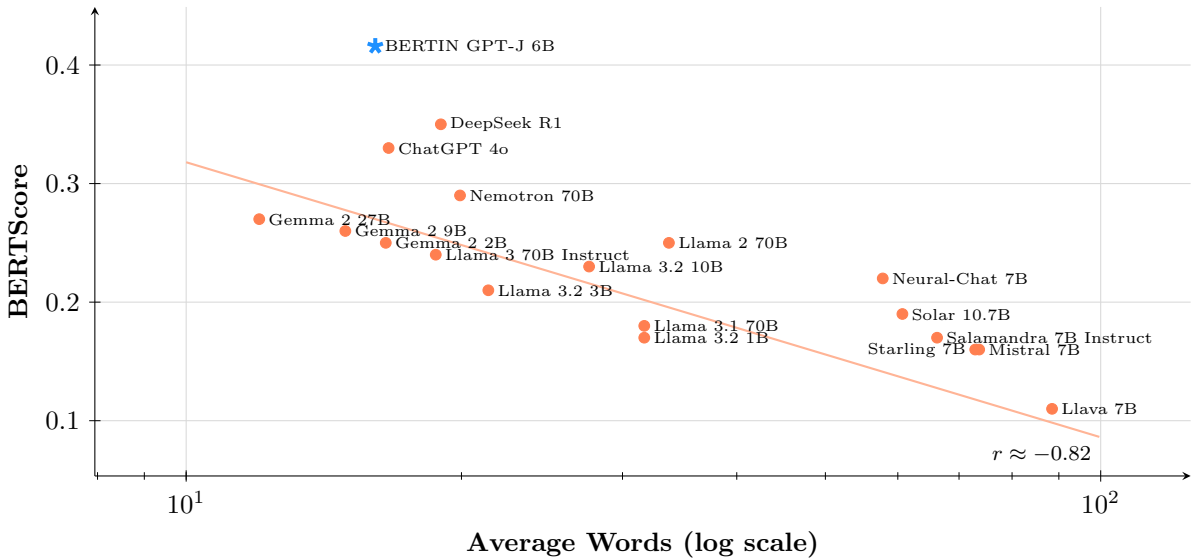
\begin{figure*}[ht]
    \small
    \centering
    \begin{tikzpicture}

\definecolor{mydotcolor}{RGB}{255,127,80} 
\definecolor{mygridcolor}{gray}{0.85}
\definecolor{mylinecolor}{RGB}{255,180,150}    
\definecolor{bertincolor}{RGB}{30,144,255} 

\begin{axis}[
    width=16cm,
    height=7.8cm,
    xlabel={\textbf{Average Words (log scale)}},
    ylabel={\textbf{BERTScore}},
    grid=major,
    grid style={mygridcolor},
    axis lines=left,
    xmode=log,
    log basis x=10,
    tick style={black},
    every tick label/.append style={font=\small},
    enlargelimits=0.1,
    legend style={font=\small, at={(0.05,0.95)}, anchor=north west}
]

\addplot[
    domain=10:100,
    samples=100,
    thick,
    color=mylinecolor
]
{-0.232*log10(x) + 0.550};

\addplot[
    only marks,
    mark=*,
    mark size=2pt,
    color=mydotcolor
] coordinates {
    (12.02, 0.27)
    (14.93, 0.26)
    (16.53, 0.25)
    (16.65, 0.33)
    (18.75, 0.24)
    (18.99, 0.35)
    (19.93, 0.29)
    (21.4, 0.21)
    (27.58, 0.23)
    (31.69, 0.18)
    (31.69, 0.17)
    (33.72, 0.25)
    (57.77, 0.22)
    (60.7, 0.19)
    (66.24, 0.17)
    (72.92, 0.16)
    (73.65, 0.16)
    (88.48, 0.11)
};

\addplot[
    only marks,
    mark=star,
    mark size=3pt,        
    line width=1.5pt,       
    color=bertincolor
] coordinates {
    (16.10, 0.416)
};

\node[anchor=west, font=\tiny] at (axis cs:12.02, 0.27) {Gemma 2 27B};
\node[anchor=west, font=\tiny] at (axis cs:14.93, 0.26) {Gemma 2 9B};
\node[anchor=west, font=\tiny] at (axis cs:16.10, 0.416) {BERTIN GPT-J 6B};
\node[anchor=west, font=\tiny] at (axis cs:16.53, 0.25) {Gemma 2 2B};
\node[anchor=west, font=\tiny] at (axis cs:16.65, 0.33) {ChatGPT 4o};
\node[anchor=west, font=\tiny] at (axis cs:18.75, 0.24) {Llama 3 70B Instruct};
\node[anchor=west, font=\tiny] at (axis cs:18.99, 0.35) {DeepSeek R1};
\node[anchor=west, font=\tiny] at (axis cs:19.93, 0.29) {Nemotron 70B};
\node[anchor=west, font=\tiny] at (axis cs:21.4, 0.21) {Llama 3.2 3B};
\node[anchor=west, font=\tiny] at (axis cs:27.58, 0.23) {Llama 3.2 10B};
\node[anchor=west, font=\tiny] at (axis cs:31.69, 0.18) {Llama 3.1 70B};
\node[anchor=west, font=\tiny] at (axis cs:31.69, 0.17) {Llama 3.2 1B};
\node[anchor=west, font=\tiny] at (axis cs:33.72, 0.25) {Llama 2 70B};
\node[anchor=west, font=\tiny] at (axis cs:57.77, 0.22) {Neural-Chat 7B};
\node[anchor=west, font=\tiny] at (axis cs:60.7, 0.19) {Solar 10.7B};
\node[anchor=west, font=\tiny] at (axis cs:66.24, 0.17) {Salamandra 7B Instruct};
\node[anchor=east, font=\tiny] at (axis cs:72.92, 0.16) {Starling 7B};
\node[anchor=west, font=\tiny] at (axis cs:73.65, 0.16) {Mistral 7B};
\node[anchor=west, font=\tiny] at (axis cs:88.48, 0.11) {Llava 7B};

\node[anchor=north east, font=\scriptsize] at (axis cs:100, {-0.232*log10(100)+0.550}) {$r \approx -0.82$};

\end{axis}
\end{tikzpicture}
    \caption{BERTScore against average number of words in generated summaries (log scale). BERTIN GPT-J 6B is highlighted with a bold blue star and excluded from the regression calculation.}
    \label{fig:wordsberts}
\end{figure*}

\section{Discussion and Future Work}
\label{s:discussion}

Given the evident lack of summarization datasets in Spanish, BOE-XSUM will enable the training of models that generate extreme summaries in clear language, as well as to categorize this type of texts. We have demonstrated experimentally that this is a complex task and a major challenge for generative models, regardless of size. The low number of summarized texts in certain categories raises the possibility of discarding them or merging them into the category \texttt{OTROS}, but we decided against because these categories are very relevant to the dataset.

Our results show that the task is more effective when fine-tuning is performed. All our fine-tuned models with precision up to 8-bit outperform models in zero-shot settings, except ChatGPT 4o and DeepSeek. It is important to notice that the fine-tuned models have only 6B parameters, in contrast with the 671B parameters of DeepSeek R1.

However, the comparison between fine-tuned and zero-shot models is not entirely symmetrical. Fine-tuned models were specifically trained on the BOE-XSUM dataset, with input-output structures tailored to the task, whereas zero-shot models were evaluated using a single static prompt, without domain-specific tuning or few-shot techniques. Nonetheless, it was exciting to discover that smaller and limited models (6B parameters, 2048-token context window) could outperform much larger frontier models when fine-tuned for a narrowly defined and linguistically constrained task. In that sense, the results showcase the power of targeted fine-tuning in legal summarization scenarios. Future work should continue exploring few-shot learning and prompt optimization for general-purpose models, in order to better understand their potential when properly guided.

Although much remains to be done, due to the complexity of the task, extreme summarization of legal texts as extensive as those in the BOE remains a significant challenge for current models. Key open questions include (i) to what extent automatic metrics are correctly assessing model behavior, and (ii) what is the relative complexity contributed by the two core aspects of the task, i.e., extreme summarization and translation to plain language.

We also acknowledge the limitations of relying on standard automatic metrics such as BLEU, ROUGE, METEOR, and BERTScore for this specific task. These metrics may fail to capture essential qualities such as clarity, usefulness, or alignment with non-specialist expectations. Notably, we observed a strong negative correlation between the length of the generated summaries and the BERTScore (-0.82), which suggests that more informative summaries may be unfairly penalized for being slightly longer. This raises concerns about whether these metrics are truly aligned with the human notion of "better" summaries in this setting. While we conducted a limited human evaluation, a more robust human-centered validation is needed.

Moreover, while the dataset originated from journalistic summaries, we emphasize that all summaries used in our experiments were carefully edited to eliminate stylistic idiosyncrasies and ensure fidelity to the BOE source. This reduces the risk of source bias and provides a more standardized and generalizable input for model training and evaluation.

Future work should thus advance in several directions. First, a robust category classifier could be developed to automatically assign thematic labels to BOE texts, enhancing the utility of the dataset and enabling more nuanced analysis. Second, the introduction of more accurate and task-specific evaluation metrics is crucial. These could include clarity-based judgments or be inspired by natural language inference or question-answering paradigms, aiming to capture the communicative goals of extreme summarization more effectively than traditional n-gram-based or embedding-based scores. Together with expanded human evaluation protocols, these lines of work will help consolidate this task as a benchmark for controlled, high-precision text generation in legal and administrative domains.
\section{Conclusions}
\label{s:conclusions}

We have developed a spanish dataset in the highly sought-after legal domain, designed to train and evaluate extreme summary generation models in clear language, suitable for microblogging platforms such as X or Bluesky. The dataset contributes to the increasingly relevant task of bridging the gap between legal language and the citizens affected by legal texts. The dataset presented will facilitate the training of systems that help the supervision of public administrations and allow the detection of relevant BOE entries that are of public interest. 

Our preliminary experiments with the dataset indicate that (even small) fine-tuned systems seem a better choice than unsupervised (prompted) frontier models. However, these results are obtained with automatic evaluation measures, and it remains to be verified whether such measures are adequate in this context. Another lines of future work involve applying prompt engineering to optimize the performance of unsupervised models, and determining the relative challenge posed by the summarization factor and the translation to clear language factor. Also, we plan to enhance the dataset with lenghtier summaries from entries in the Civio website. 

Finally, given that we have a categorized dataset with extreme summaries of articles published in the BOE, future work could focus on leveraging this dataset to develop a model capable of classifying all daily BOE entries. A generative model could be trained to automatically generate summaries for the classified entries. This would enable the automation and publication of extreme-summaries for BOE entries that are of broad public interest.

\section*{Acknowledgments}
We thank David Cabo, co-director and CTO of the Civio Foundation, for his continued efforts to promote transparency in public institutions. We are especially grateful to co-director and journalist Eva Belmonte for her outstanding work making the Spanish Official State Gazette (BOE) understandable and relevant to the public. Through her project ``El BOE nuestro de cada día," (\textit{``Our daily BOE"}) she has highlighted the Gazette's most impactful content, helping bridge the gap between government actions and citizen awareness.

This work has been partially funded by the European Union - NextGenerationEU through the `Recovery, Transformation and Resilience Plan', by the Ministry of Economic Affairs and Digital Transformation. Additionally, we thank Google for providing compute resources via the Tensor Research Cloud program, which significantly supported our model training efforts.

\bibliographystyle{fullname}
\bibliography{references}

\newpage
\onecolumn
\appendix


\section{Editor server}
\label{ap:editor}

\begin{figure*}[h]
    \centering
    \includegraphics[width=\textwidth]{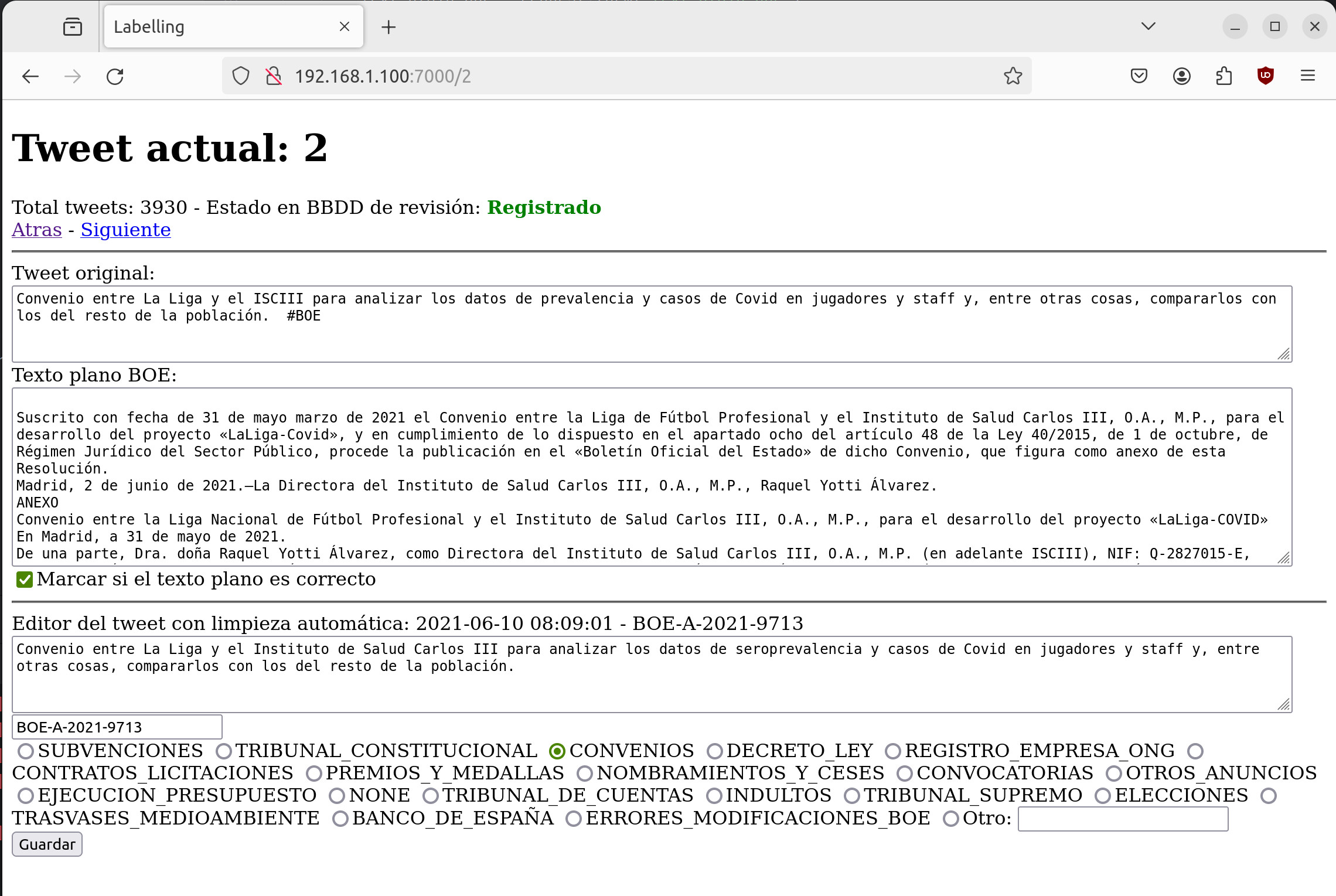}
    \caption{Editor server.
    \label{fig:server_editor}}
    
\end{figure*}

 The features of this editing server include: 

\begin{itemize}
\item Display of the current tweet number, facilitating easy navigation and reference.
\item Indication of the data item's status within the new database, distinguishing between {Registered and Not registered} statuses.
\item Navigation functionality, including Back and Forward buttons, which allow users to seamlessly move through the dataset.
\item Inclusion of the original tweet authored by Eva Belmonte, providing a direct link to the initial data point.
\item Presentation of the plain text version of the BOE documents, essential for content verification and analysis.
\item A verification checkbox to confirm the accuracy of the BOE's plain text, addressing instances of incorrect document linkage. Despite the rarity of such occurrences, all affected entries have been excluded from the dataset.
\item An automated cleanup editor for the tweet, designed to enhance data cleanliness and usability.
\item The identifier of the BOE document, enabling users to verify the relevance and accuracy of the associated tweet.
\item Categorization functionality for each data item, facilitating structured analysis and retrieval.
\item A Save button to ensure any modifications or tags applied are retained.
\end{itemize}

\newpage

\section{Categories annotated for the text categorization task}
\label{ap:catgories}

\begin{table*}[ht]
\footnotesize
\centering
\begin{tabular}{|c|>{\raggedright\arraybackslash}p{10cm}|}
\hline
\textbf{Category} & \textbf{Examples} \\
\hline
\texttt{OTROS\_ANUNCIOS} & \parbox[t]{10cm}{\textit{El Gobierno publica la nueva metodología de cálculo precios pequeño consumidor energía}} \\
\hline
\texttt{CONTRATOS\_LICITACIONES} & \parbox[t]{10cm}{\textit{La readjudicación del contrato a Aguas de Valencia por importe total de 7.000.000 euros}} \\
\hline
\texttt{NOMBRAMIENTOS\_Y\_CESES} & \parbox[t]{10cm}{\textit{Nombrada comisionada del gobierno frente al reto demográfico a Edelmira Barreira Diz.}} \\
\hline
\texttt{TRIBUNAL\_CONSTITUCIONAL} & \parbox[t]{10cm}{\textit{Sentencia del TC que anula los límites de Cataluña a la libertad de horarios comerciales.}} \\
\hline
\texttt{SUBVENCIONES} & \parbox[t]{10cm}{\textit{Subvenciones a sindicatos del Ministerio de Empleo: 8.883.890 euros.}} \\
\hline
\texttt{DECRETO\_LEY} & \parbox[t]{10cm}{\textit{Ley de facturación electrónica. Para contratos con la administración: todos obligados a partir de 15 enero 2015}} \\
\hline
\texttt{TRIBUNAL\_DE\_CUENTAS} & \parbox[t]{10cm}{\textit{El Tribunal de Cuentas busca más de 400.000 euros perdidos en liquidación en la Cámara Oficial de la Propiedad Urbana de Vizcaya}} \\
\hline
\texttt{CONVOCATORIAS} & \parbox[t]{10cm}{\textit{El Ministro de Hacienda y Administraciones Públicas ha dispuesto hacer pública la adjudicación de puestos de trabajo especificados en el anexo a la presente orden.}} \\
\hline
\texttt{PREMIOS\_Y\_MEDALLAS} & \parbox[t]{10cm}{\textit{Premio Nacional de Historia de España 2013 a José Ángel Sánchez Asiain}} \\
\hline
\texttt{CONVENIOS} & \parbox[t]{10cm}{\textit{Ratificación y próxima entrada en vigor (el 1 de abril) del convenio de doble nacionalidad Francia-España firmado hace un año. Permite tener ambas nacionalidades (también recuperarla si se perdió para tener la del otro país)}} \\
\hline
\texttt{TRIBUNAL\_SUPREMO} & \parbox[t]{10cm}{\textit{Karl Friedrich Schober recurre contra la Orden HAP/72/2013, la declaración informativa sobre bienes y derechos situados en el extranjero, ante la Audiencia Nacional}} \\
\hline
\texttt{REGISTRO\_EMPRESA\_ONG} & \parbox[t]{10cm}{\textit{Exteriores crea una Oficina Consular Honoraria en Incheon (Corea del Sur) para relaciones económicas.}} \\
\hline
\texttt{ERRORES\_MODIFICACIONES\_BOE} & \parbox[t]{10cm}{\textit{Corrección de errores para quitar la marca en la casilla ‘trabajador sin especialización’ sobre la madre}} \\
\hline
\texttt{ELECCIONES} & \parbox[t]{10cm}{\textit{Resultados definitivos elecciones municipales, hasta la J de Jaen.}} \\
\hline
\texttt{TRASVASES\_MEDIOAMBIENTE} & \parbox[t]{10cm}{\textit{La ley de contaminación y residuos que entra en vigor mañana. Acorta plazo para conseguir autorización ambiental}} \\
\hline
\texttt{EJECUCION\_PRESUPUESTO} & \parbox[t]{10cm}{\textit{Ejecución del presupuesto en junio.}} \\
\hline
\texttt{INDULTOS} & \parbox[t]{10cm}{\textit{Indulto a María Salmerón Parrilla aprobado viernes en Consejo de Ministros.}} \\
\hline
\texttt{BANCO\_DE\_ESPAÑA} & \parbox[t]{10cm}{\textit{Multa del Banco de España a Austrogiros Entidad de Pago, S.A. por un reguero de incumplimientos de la ley (no tener dirección en España, irregularidades contables, etc...): 1.300.000 euros más multas a los dos administradores e inhabilitación.}} \\
\hline
\end{tabular}
\caption{Categories annotated for the text categorization task and examples.} \label{t:categories-examples}
\end{table*}

\begin{table*}[htb]
\footnotesize
\centering
\begin{tabular}{lrrrr}
\textbf{Category} & \textbf{All} & \textbf{Train} & \textbf{Development} &  \textbf{Test} \\
\midrule
OTROS\_ANUNCIOS 	&	 1004 &	 789 &	 111 &	 104 \\
CONTRATOS\_LICITACIONES 	&	 648 &	 505 &	 73 &	 70 \\
NOMBRAMIENTOS\_Y\_CESES 	&	 324 &	 255 &	 35 &	 35  \\
TRIBUNAL\_CONSTITUCIONAL 	&	 311 &	 245 &	 31 &	 34 \\
SUBVENCIONES 	&	 228 &	 177 &	 25 &	 26 \\
DECRETO\_LEY 	&	 173 &	 137 &	 19 &	 17 \\
TRIBUNAL\_DE\_CUENTAS 	&	 161 &	 128 &	 18 &	 17 \\
CONVOCATORIAS 	&	 155 &	 120 &	 17 &	 17 \\
PREMIOS\_Y\_MEDALLAS 	&	 152 &	 119 &	 16 &	 16 \\
CONVENIOS 	&	 141 &	 112 &	 12 &	 15 \\
TRIBUNAL\_SUPREMO 	&	 97 &	 75 &	 11 &	 11 \\
REGISTRO\_EMPRESA\_ONG 	&	 82 &	 63 &	 9 &	 10 \\
ERRORES\_MODIFICACIONES\_BOE 	&	 52 &	 44 &	 4 &	 5 \\
ELECCIONES 	&	 37 &	 30 &	 3 &	 4 \\
TRASVASES\_MEDIOAMBIENTE 	&	 33 &	 27 &	 3 &	 2 \\
EJECUCION\_PRESUPUESTO 	&	 25 &	 21 &	 2 &	 2 \\
INDULTOS 	&	 20 &	 16 &	 2 &	 2 \\
BANCO\_DE\_ESPAÑA 	&	 5 & 	 4 & 	 1 & 0	\\
\end{tabular}
\caption{Distribution of text categories annotated in BOE-XSUM.} \label{t:distribution-categories}
\end{table*}


\section{Additional Information about the Dataset}
\label{ap:additional-info-dataset}

\begin{table*}[htb]
\centering
\small
\begin{tabular}{lp{10cm}}
\textbf{Column} & \textbf{Description} \\
\midrule
id & Unique item identifier. \\
boe\_materials & BOE category identifier. \\
boe\_date\_publication & Publication date of the BOE article. \\
boe\_previous & Previous BOE articles that are modified by this new BOE. \\
boe\_id & BOE identifier. \\
boe\_title & Title of the BOE article. \\
boe\_soup\_xml & Complete scraped web page. \\
tweet\_original & Original tweet by Eva Belmonte. \\
boe\_category & Category to which this item belongs. \\
boe\_alert & BOE classification codes for government areas. \\
boe\_departament & Government department that issued the BOE article. \\
tweet\_text\_cleaned & Extreme summary generated from a thorough review of Eva Belmonte's tweet. \\
boe\_subsequent & Subsequent legislation articles modified by this order (Only for articles referring to laws). \\
\end{tabular}
\caption{Dataset Columns.}
\label{tab:dataset_columns}
\end{table*}

\FloatBarrier


\begin{table}[htb]
\centering
\footnotesize
\begin{tabular}{@{}lr|lr@{}}
\textbf{Range} & \textbf{Count} & \textbf{Range} & \textbf{Count}\\ \midrule
90\%+ &	1154  &	40\%+  & 2946 \\
80\%+  & 1624 &	30\%+  & 3162 \\
70\%+  & 2028 &	20\%+  & 3341 \\
60\%+  & 2398 &	10\%+ &	 3482 \\
50\%+  & 2720 &	0\%+  &	 3648 \\ 
\end{tabular}
\caption{Distribution of cosine similarity ranges for the pairs of original and modified tweets. \label{t:similarity-distribution}}
\end{table}

\begin{table*}[htb]
\small
\begin{tabular}{lrrrrrr}

      & {\bf \# pairs} & {\bf\# words } & {\bf\# words} & {\bf average \#} & {\bf average \#} & {\bf compression} \\ 
     & {\bf doc/sum } & {\bf doc} & {\bf sum} & {\bf  words/doc} & {\bf words/sum} & {\bf rate} \\ \midrule
Train & 2,867        & 10,859,884        & 49,461              & 3,787 & 17 & 0.0045\% \\ 
Dev   & 392         & 1,155,745         & 6,733               & 2,948 & 17 & 0.0058\% \\ 
Test  & 389         & 1,367,413         & 6,605               & 3,515 & 17 & 0.0048\%\\ 
{\bf Total} & 3,648        & 13,383,042        & 62,799              & 3,668 & 17 & 0.0050\%\\
\end{tabular}
\caption{Dataset statistics showing  the total number of items, word counts in documents and summaries, mean word count in documents and summaries, and the compression ratio for the train, development, and test sets.\label{tab:dataset_stats}}
\end{table*}


\section{Examples of edited summaries}

\label{ap:edited}

As an example, we edited the summary below because the content of the post gave very little information about the BOE article it referred to:

{\small
\begin{examples}
\item {\bf Original}: \textit{``Es habitual tratar el tema Microsoft en administraciones públicas, pero solemos olvidar la presencia de Oracle \#BOE''}\\
 {\bf Edited}: \textit{``La Agencia Estatal de Seguridad Aérea contrata a Oracle por el servicio de mantenimiento y soporte técnico por un importe total de 1.232.070,40 euros.''} \label{ex:oracle-editado}
\item {\bf Original}: \textit{``It is common to talk about Microsoft in public administrations, but we tend to forget the presence of Oracle.''}\\
{\bf Edited}: \textit{``The Aviation Safety State Agency contracts Oracle for the maintenance and technical support service for a total amount of 1,232,070.40 euros.''} \label{ex:oracle-editado}
\end{examples}
}

In the next example, we edited the content because the post was talking about people and entities not present in the BOE article.

{\small
\begin{examples}
\item {\bf Original}: \textit{``La Moncloa tiene que licitar su servicio de restauración de forma urgente porque la empresa anterior que daba este servicio, Dulcinea nutrición -que hasta 2018 presidía uno de los bisnietos de Franco y ahora está en concurso-, dejó de pagar impuestos. \#BOE"}.\\
 {\bf Edited}: \textit{``Licitación del servicio de restauración, de limpieza de los espacios destinados a dicha finalidad y de máquinas de venta automática en el Complejo de La Moncloa."}.
\item {\bf Original}: \textit{``La Moncloa has to tender its catering service urgently because the previous company that provided this service, Dulcinea nutrición - which until 2018 was presided over by one of Franco's great-grandchildren and is now in competition - stopped paying taxes.. \#BOE"}.\\
 {\bf Edited}: \textit{``Bidding for catering services, cleaning of the areas destined for this purpose and vending machines in the Moncloa Complex."}.
\end{examples}
}

\section{Additional Information about Results}
\label{ap:additional-results}

\begin{table*}[htb]
\centering
\footnotesize
\begin{tabular}{lcccc}
\textbf{Model} & \textbf{BLEU} & \textbf{METEOR} & \textbf{ROUGE} & \textbf{BERTScore} \\
\midrule
Llama 3.2 1B & 0.010 & 0.137 & 0.171 & 0.124 \\
Llama 3.1 8B & 0.015 & 0.185 & 0.220 & 0.196 \\
Llama 3.1 70B & 0.016 & \textbf{0.220} & 0.214 & \textbf{0.207} \\
Llama 3 70B Instruct & 0.017 & 0.175 & 0.204 & 0.178 \\
Llama 2 70B & 0.009 & 0.110 & 0.154 & 0.117 \\
Gemma 2 2B & 0.015 & 0.183 & 0.211 & 0.198 \\
Gemma 2 9B & \textbf{0.019} & 0.207 & 0.232 & 0.213 \\
Gemma 2 27B & 0.018 & 0.193 & 0.214 & 0.197 \\
Mistral 7B & 0.006 & 0.094 & 0.152 & 0.106 \\
Neural-Chat 7B & 0.006 & 0.110 & 0.167 & 0.132 \\
Starling 7B & 0.003 & 0.074 & 0.134 & 0.092 \\
Llava 7B & 0.002 & 0.063 & 0.114 & 0.069 \\
Solar 10.7B & 0.005 & 0.102 & 0.159 & 0.118 \\
Nemotron 70B & 0.012 & 0.143 & 0.185 & 0.132 \\
Salamandra 7B Instruct & 0.004 & 0.082 & 0.127 & 0.070 \\
ChatGPT 4o & \textbf{0.019} & 0.208 & \textbf{0.236} & 0.206 \\
\end{tabular}
\caption{Results on test set with unmodified social media posts by the journalist. Best scores in \textbf{bold}.
\label{tab:baseline_sim_1}
}
\end{table*}

\begin{figure*}[ht]
    \centering
    \includegraphics[width=0.8\textwidth]{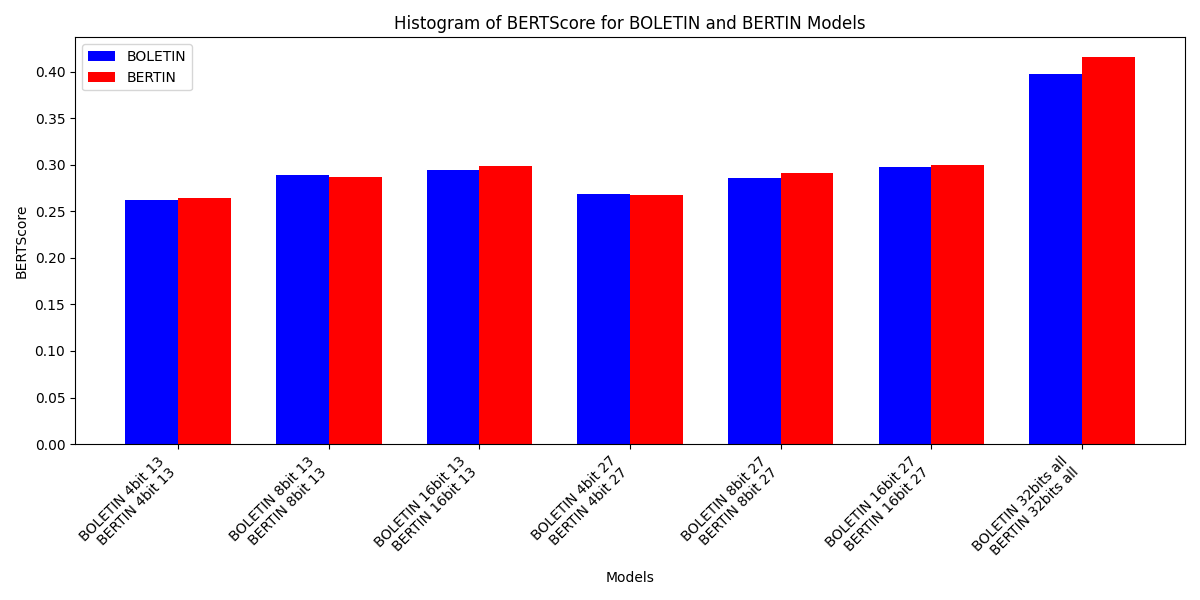}  
    \caption{Comparative results between Boletin and Bertin across bit widths and training configurations.
    \label{fig:comparative_boletin_bertin}
    } 
\end{figure*}

\begin{table*}[htb]
\centering
\footnotesize
\begin{tabularx}{\textwidth}{|X|c|X|c|X|c|}
\hline
\textbf{Summary Gemma 2 9B} & \textbf{F1} & \textbf{Summary LLama 2 70B } & \textbf{F1} & \textbf{Tweet} & \textbf{Dif} \\ \hline
José Antonio Pérez Tapias es nombrado Catedrático de Filosofía en la Universidad de Granada. & 0.79 & La Rectora de la Universidad de Granada, María Pilar Aranda Ramírez, ha nombrado a José Antonio Pérez Tapias como Catedrático de Universidad del área de conocimiento de Filosofía, adscrito al Departamento de Filosofía II, mediante Resolución de 18 de enero de 2017. & 0.38 & Pérez Tapias ya es Catedrático de Filosofía en la Universidad de Granada. & 0.41 \\ \hline
Se licita el mantenimiento y soporte técnico de sistemas Oracle de la Agencia Estatal de Seguridad Aérea por valor estimado de 1.018.240 euros. & 0.63 & El presidente de la mesa de contratación, Eusebio Jiménez Arroyo, anunció que se busca una empresa para el mantenimiento y soporte técnico de los sistemas de información en tecnología Oracle de la Agencia Estatal de Seguridad Aérea. El contrato tendrá una duración de 20 meses y se adjudicará a la empresa que ofrezca los mejores criterios de adjudicación. La fecha límite para la presentación de ofertas es el 3 de junio de 2014. & 0.19 & La Agencia Estatal de Seguridad Aérea contrata a Oracle por el servicio de mantenimiento y soporte técnico por un importe total de 1.232.070,40 euros. & 0.44 \\ \hline
El Ayuntamiento de Barcelona licita un sistema para controlar infracciones medioambientales de vehículos en la zona de bajas emisiones. & 0.66 & El Ayuntamiento de Barcelona invita a licitar para un contrato de suministro e instalación de un sistema de monitorización y control de infracciones medioambientales en la ciudad. El plazo de ejecución es de 11 meses, y se requiere una garantía definitiva del 5\% del ... (continue) & 0.12 & Barcelona licita el contrato para el control de los vehículos en la zona de bajas emisiones. & 0.54 \\ \hline
\end{tabularx}
\caption{Table with summaries, F1 scores, and differences.
\label{tab:example_1_gemma2vsLlama2}
}
\end{table*}


\FloatBarrier

\section{English Version of BOE example with original and edited summaries}\label{app:english_example}
\begin{mdframed}
\footnotesize
Royal Decree-Law 6/2012, of March 9, on urgent measures for the protection of mortgage debtors without resources, establishes a series of mechanisms aimed at enabling the restructuring of mortgage debt for those experiencing extraordinary difficulties in making their payments.  

To this end, the aforementioned Royal Decree-Law incorporated a Code of Good Practices, to which entities may adhere, and whose compliance will be supervised by a control commission. The composition of this commission was modified by Article 6 of Law 1/2013, of May 14, on measures for...  

... – Caja Rural San Jaime de Alquerías Niño Perdido, S. Coop. de Crédito V.  
– Caja Rural San José de Almassora, S. Coop. de Crédito V.  
– Caja Rural San José de Burriana, S. Coop. de Crédito V.  
– Caja Rural San José de Nules, S. Coop. de Crédito V.  
– Caja Rural San Roque de Almenara, S. Coop. de Crédito V.  
– Colonya–Caixa d’Estalvis de Pollença  
– Liberbank, S.A.  
– Publicredit, S.L.  
– UNOE Bank, S.A.  
\end{mdframed}

The original summary for this BOE article as written by the journalist is as follows:

\begin{mdframed}
\footnotesize
Adhesions to 2 codes of good practices in mortgages. So far, not very effective. Because legislating is hard, right? \#BOE
\end{mdframed}

The resulting edited summary:

\begin{mdframed}
\footnotesize
List of bank adhesions to the codes of good practices to strengthen the protection of mortgage debtors, debt restructuring, and social rent
\end{mdframed}


\section{Annotation Guidelines}\label{app:guidelines}

\small
To improve consistency, objectivity, and reliability, these annotation guidelines will help the annotators create concise, accurate, and legally faithful extreme summaries of BOE texts.

\textbf{1. General Principles}

Each summary must:
\begin{itemize}  
    \item Be concise -- Between 15 and 22 words.
    \item Be factual -- Avoid opinions, subjective language, or speculation.
    \item Preserve critical legal information -- Names, amounts, dates, and key legal actions must be included.
    \item Avoid redundancy -- Keep the summary clear and direct.
    \item Use simple but precise language -- The goal is to make legal information more accessible.
\end{itemize}

\textbf{2. Formatting Rules}

\begin{itemize}
    \item Use neutral, third-person language (No opinions, interpretations, or assumptions).
    \item Avoid legal jargon unless necessary -- If a legal term is essential, keep it. If it can be simplified without changing meaning, do so.
    \item No abbreviations unless widely known (e.g., BOE, EU, UN are fine).
\end{itemize}

\textbf{Bad Example} \ding{55} (Too vague \& informal)

\begin{quote}
``Another BOE article about a contract for an unnamed company.''
\end{quote}

\textbf{Good Example} \ding{52} (Clear, factual, and legally informative)

\begin{quote}
``The Ministry of Transport awards a \euro{2M} contract to Ferrovial for highway maintenance in Madrid.''
\end{quote}

\textbf{3. Information Hierarchy}

\textit{What to include first? If space is limited, prioritize information in this order:}

\begin{tabular}{l | m{5cm} | m{7cm}}
Priority & Include in Summary & Example \\
\hline
\ding{172} & Main Legal Action & ``The government approves a new tax reduction for self-employed workers.'' \\
\ding{173} & Who is involved? & ``The Ministry of Defense signs a \euro{5M} cybersecurity contract with IBM.'' \\
\ding{174} & Financial or Legal Consequences & ``A judge orders a company to pay \euro{1.2M} for contract fraud.'' \\
\ding{175} & Date or Location (if critical) & ``A new rental law takes effect in Barcelona on January 1, 2025.'' \\
\end{tabular}

\vspace{1em}
If space is limited, drop information from the bottom of the list first.

\textbf{4. Common Errors \& How to Avoid Them}

\textbf{Subjectivity or Opinions}

\begin{itemize}
    \item[\ding{55}] ``A useless new law that won’t change anything.''
    \item[\ding{52}] ``The government introduces a new environmental protection law.''
\end{itemize}

\textbf{Missing Critical Information}

\begin{itemize}
    \item[\ding{55}] ``A contract was signed.'' (Who signed it? What for?)
    \item[\ding{52}] ``The Ministry of Health awards a \euro{3.5M} contract for hospital equipment.''
\end{itemize}

\textbf{Overly Complex Legal Language}

\begin{itemize}
    \item[\ding{55}] ``In accordance with Royal Decree 1892/2024, a framework agreement has been instituted for the execution of infrastructural oversight.''
    \item[\ding{52}] ``The government approves a contract for highway inspections.''
\end{itemize}

\textbf{Misleading Summaries}

\begin{itemize}
    \item[\ding{55}] ``The Supreme Court supports corruption!'' (Misinterpretation)
    \item[\ding{52}] ``The Supreme Court rules against a corruption conviction in X case.''
\end{itemize}

\textbf{5. Handling Difficult Cases}

\textbf{A) Summarizing Lengthy Legal Texts}
\begin{itemize}
    \item If a BOE article is very long, focus on the main action in the first few paragraphs.
    \item \textbf{BOE Text (Full Version)}: \\
    ``The government issues a new decree modifying Article 10 of the Labor Code, changing workplace safety regulations.''
    \item \textbf{Summary}: \\
    ``New decree modifies workplace safety regulations in Spain.''
\end{itemize}

\textbf{B) Appointments \& Dismissals}
\begin{itemize}
    \item Format: ``[Person] appointed/dismissed as [position] at [institution].''
    \item \textbf{BOE Text}: ``By Royal Decree 2024/317, José Pérez is appointed Minister of Finance.''
    \item \textbf{Summary}: ``José Pérez appointed Minister of Finance by Royal Decree.''
\end{itemize}

\textbf{C) Court Rulings}
\begin{itemize}
    \item Mention court name, ruling outcome, and key details.
    \item \textbf{BOE Text}: ``The Constitutional Court overturns a law restricting media freedom.''
    \item \textbf{Summary}: ``The Constitutional Court annuls media restriction law.''
\end{itemize}

\textbf{D) Large Financial Figures}
\begin{itemize}
    \item If space is limited, round numbers where possible.
    \item \textbf{Example}: \\
    ``The Ministry of Transport signs a \euro{2,345,678.90} contract for roads.'' \\
    Can be summarized as: \\
    ``\euro{2.3M} contract for roadworks by Ministry of Transport.''
\end{itemize}

\textbf{5. Final Checklist for Annotators}
\begin{itemize}  
    \item Did you keep the summary between 15--22 words?
    \item Is the summary neutral and free of opinion?
    \item Does it capture the key legal action (law, contract, ruling, appointment)?
    \item Does it include important details (names, amounts, institutions)?
    \item Did you avoid excessive jargon?
    \item Would a non-lawyer understand it?
\end{itemize}

\end{document}